%% file: main.tex
\documentclass[11pt,a4paper]{article}
\input{init.tex}

\title{The Future is not One-dimensional: Complex Event Schema Induction by \\ Graph Modeling for Event Prediction}

\author{Manling Li\textsuperscript{\textnormal{1}}, Sha Li\textsuperscript{\textnormal{1}}, Zhenhailong Wang\textsuperscript{\textnormal{1}}, Lifu Huang\textsuperscript{\textnormal{2}},  \\
 \textbf{Kyunghyun Cho\textsuperscript{\textnormal{3}}, Heng Ji\textsuperscript{\textnormal{1}}, Jiawei Han\textsuperscript{\textnormal{1}}, Clare Voss\textsuperscript{\textnormal{4}}} \\
  \textsuperscript{1}UIUC  
  \textsuperscript{2}Virginia Tech
  \textsuperscript{3}NYU
  \textsuperscript{4}US ARL\\
  \texttt{\fontfamily{pcr}\selectfont\{manling2,shal2,wangz3,hengji,hanj\}@illinois.edu}, \\
  \texttt{\fontfamily{pcr}\selectfont{lifuh@vt.edu}},
  \texttt{\fontfamily{pcr}\selectfont{kyunghyun.cho@nyu.edu}},
  \texttt{\fontfamily{pcr}\selectfont{clare.r.voss.civ@mail.mil}}\\
  }

\date{}

\begin{document}

\maketitle

\input{0_abstract.tex}

\input{1_introduction.tex}

\input{2_problem}

\input{3_method}

\input{4_benchmark}

\input{5_analysis}

\input{6_relatedwork.tex}

\input{7_conclusion.tex}

\section*{Acknowledgement}
This research is based upon work supported by U.S. DARPA KAIROS Program Nos. FA8750-19-2-1004 and Air Force No. FA8650-17-C-7715. The views and conclusions contained herein are those of the authors and should not be interpreted as necessarily representing the official policies, either expressed or implied, of DARPA, or the U.S. Government. The U.S. Government is authorized to reproduce and distribute reprints for governmental purposes notwithstanding any copyright annotation therein.

\bibliography{ref}
\bibliographystyle{acl_natbib}

\end{document}

%% file: init.tex
\usepackage[table,dvipsnames]{xcolor}
\usepackage{emnlp2021} %
\usepackage{times}
\usepackage{latexsym}

\usepackage{microtype}

\usepackage{graphicx}               %
\usepackage{tabularx}               %
\newcolumntype{C}{>{\centering\arraybackslash}X}
\usepackage{multirow}               %
\usepackage{diagbox}                %
\usepackage{hhline}                 %
\usepackage{color}                  %
\usepackage{amsmath}                %
\usepackage{amssymb}                %
\usepackage{mathtools}              %
\usepackage{enumitem}               %
\newcommand{\bs}{\boldsymbol}

\usepackage{subfigure}
\usepackage{booktabs}
\usepackage{extarrows}
\usepackage{makecell}

\newtheorem{thm:def}{Definition}
\newtheorem{thm:eg}{Example}
\newtheorem{thm:lem}{Lemma}

\definecolor{RoseQuartzBg}{HTML}{F7CAC9}
\definecolor{RoseQuartz}{HTML}{F5A798}
\definecolor{Serenity}{HTML}{92A8D1}
\definecolor{OrangeRed}{rgb}{1.0, 0.27, 0.0}
\definecolor{Turquoise}{HTML}{0F4C81}
\definecolor{mint}{rgb}{0.24, 0.71, 0.54}
\definecolor{green}{rgb}{0.0, 0.120, 0.0}

\usepackage{xparse}
\NewDocumentCommand{\heng}{ mO{} }{\textcolor{OrangeRed}{\textsuperscript{\textit{Heng}}\textsf{\textbf{\small[#1]}}}}
\NewDocumentCommand{\kyunghyun}{ mO{} }{\textcolor{Serenity}{\textsuperscript{\textit{Kyunghyun}}\textsf{\textbf{\small[#1]}}}}
\NewDocumentCommand{\lifu}{ mO{} }{\textcolor{Turquoise}{\textsuperscript{\textit{Lifu}}\textsf{\textbf{\small[#1]}}}}
\NewDocumentCommand{\vicki}{ mO{} }{\textcolor{red}{\textsuperscript{\textit{Vicki}}\textsf{\textbf{\small[#1]}}}}
\NewDocumentCommand{\zoey}{ mO{} }{}
\NewDocumentCommand{\manling}{ mO{} }{\textcolor{red}{\textsuperscript{\textit{Manling}}\textsf{\textbf{\small[#1]}}}}
\NewDocumentCommand{\zhenhailong}{ mO{} }{\textcolor{red}{\textsuperscript{\textit{zhenhailong}}\textsf{\textbf{\small[#1]}}}}
\NewDocumentCommand{\jh}{ mO{} }{\textcolor{purple}{\textsuperscript{\textit{JW}}\textsf{\textbf{\small[#1]}}}}
\NewDocumentCommand{\cv}{ mO{} }{\textcolor{blue}{\textsuperscript{\textit{Clare}}\textsf{\textbf{\small[#1]}}}}

%% file: 0_abstract.tex
\begin{abstract}

Event schemas encode knowledge of stereotypical structures of events and their connections. As events unfold, schemas are crucial to act as a scaffolding. Previous work on event schema induction focuses either on atomic events or linear temporal event sequences, ignoring the interplay between events via arguments and argument relations. We introduce a new concept of \textit{Temporal Complex Event Schema}: a graph-based schema representation that encompasses events, arguments, temporal connections and argument relations. In addition, we propose a \textit{Temporal Event Graph Model} that predicts event instances following the temporal complex event schema. To build and evaluate such schemas, we release a new schema learning corpus containing 6,399 documents accompanied with event graphs, and we have manually constructed gold-standard schemas. Intrinsic evaluations by \textit{schema matching} and \textit{instance graph perplexity}, prove the superior quality of our probabilistic graph schema library compared to linear representations. Extrinsic evaluation on \textit{schema-guided future event prediction} further demonstrates the predictive power of our event graph model, significantly outperforming human schemas and baselines by more than 23.8\% on HITS@1. \footnote{The programs, data and resources are made publicly available for research purpose in \url{https://github.com/limanling/temporal-graph-schema}.}
\end{abstract}

%% file: 1_introduction.tex
\section{Introduction}
The current automated event understanding task has been overly simplified to be local and sequential. Real world events, such as disease outbreaks and terrorist attacks, have multiple actors, complex timelines, intertwined relations and multiple possible outcomes.
Understanding such events requires knowledge in the form of a library of event schemas, capturing the progress of time, and performing global inference for event prediction. 
For example, regarding the 2019 protest in Hong Kong International Airport, a typical question from analysts would be ``{\em How long will the flights being canceled?}" This requires an event understanding system to match events to schema representations and reason about what might happen next. The {\it airport protest} schema would be triggered by ``protest'' and ``flight cancellation'', and evidence of protesters (e.g., the number of protesters, the instruments being used, etc) will suggest a CEO resignation event, or a flight rescheduling event, or continuous flight cancellation events with respective probabilities. %

\begin{figure}[h]
    \centering
    \includegraphics[width=1.0\linewidth]{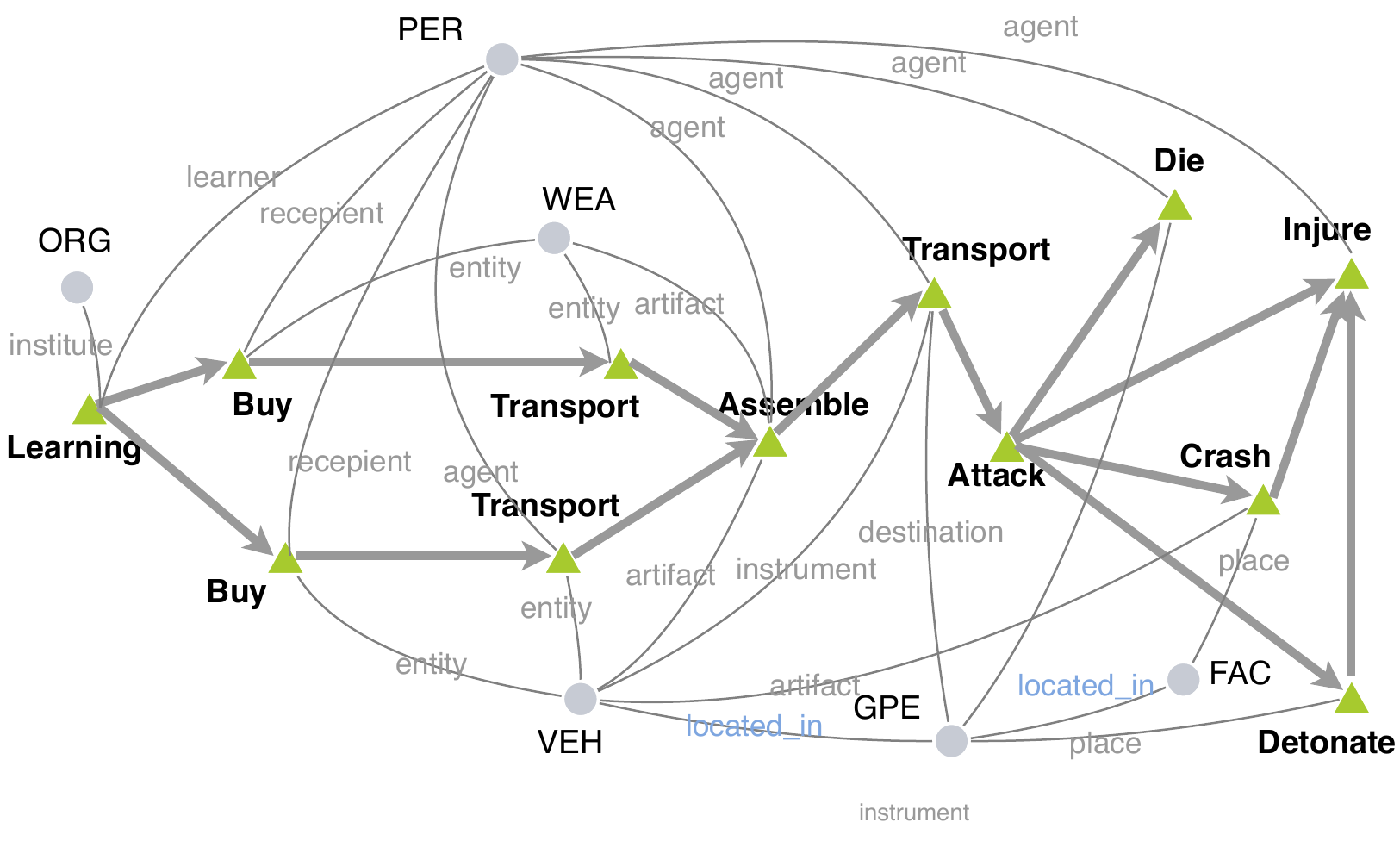}
    \caption{The example schema of the complex event type \textit{car-bombing}.  %
    A person learned to make bombs and bought materials as well as a vehicle. Then the bomb was assembled to the vehicle, and then the attacker drove it to attack people. People can be hurt by the vehicle, or by the explosion of the bomb, or by the crash of the vehicle. 
    }
    \label{fig:schema-example}
\end{figure}

Comprehending such a news story requires following a timeline, identifying key events and tracking characters. We refer to such a ``story'' as a \textbf{complex event}, e.g., the \textit{Kabul ambulance bombing} event. Its complexity comes from the inclusion of multiple atomic events (and their arguments), relations and temporal order. 
A \textbf{complex event schema} can be used to define the typical structure of a particular type of {complex event}, e.g., \textit{car-bombing}. %
This leads us to the new task that we address in this paper: \textbf{temporal complex event schema induction}.
\figurename~\ref{fig:schema-example} shows an example schema about \textit{car-bombing} %
with multiple temporal dependencies between events. Namely, the occurrence of one event may depend on multiple events. For example, the \textsc{Assemble} event happens after buying both the bomb materials and the vehicle. Also, there may be multiple events following an event, such as the multiple consequences of the \textsc{Attack} event in \figurename~\ref{fig:schema-example}. That is to say, ``{the future is not one-dimensional}''. Our automatically induced probabilistic complex event schema can be used to forecast event abstractions into the future and thus provide a comprehensive understanding of evolving situations, events, and trends.

For each type of complex event, we aim to induce a schema library that is probabilistic, temporally organized and semantically coherent. Low level atomic event schemas %
are abundant, and can be part of multiple, sparsely occurring, higher-level schemas. We propose a \textbf{Temporal Event Graph Model}, an auto-regressive graph generation model, to reach this goal.
Given a currently extracted event graph, we generate the next event type node with its potential arguments, such as the \textsc{Arrest} event in \figurename~\ref{fig:overview}, and then propagate edge-aware information following temporal orders. 
After that, we employ a copy mechanism to generate coreferential arguments, such as the \textsc{Detainee} argument is the \textsc{Attacker} of the previous \textsc{Attack} event, and build relation edges for them, e.g., \textsc{part\_whole} relation between the \textsc{Place} arguments. Finally, temporal dependencies are determined with argument connections considered, such as the temporal edge showing that \textsc{Arrest} is after \textsc{Attack}. %

Our generative model serves as both a schema library and a predictive model. Specifically, we can probe the model to generate event graphs unconditionally to obtain a set of schemas. We can also pass partially instantiated graphs to the model and ``grow'' the graph either forward or backward in time to predict missing events, arguments or relations, both from the past and in the future. 
We propose a set of schema matching metrics to evaluate the induced schemas by comparing with human-created schemas and show the power of the probabilistic schema in %
the task of \textit{future event prediction} as an extrinsic evaluation, to predict event types that are likely to happen next.

We make the following novel contributions:
\begin{itemize}[leftmargin=*]\setlength\itemsep{-0.4em}
\item This is the first work to induce probabilistic temporal graph schemas for complex events across documents, which capture temporal dynamics and connections among individual events through their coreferential or related arguments. 

\item This is the first application of graph generation methods to induce event schemas.

\item This is the first work to use complex event schemas for event type prediction, and also produce multiple hypotheses with probabilities. 

\item We have proposed a comprehensive set of metrics for both intrinsic and extrinsic evaluations.

\item We release a new data set of 6,399 documents with gold-standard schemas annotated manually. %

\end{itemize}

%% file: 2_problem.tex
\begin{figure*}[h]
    \centering
    \includegraphics[width=1.0\linewidth]{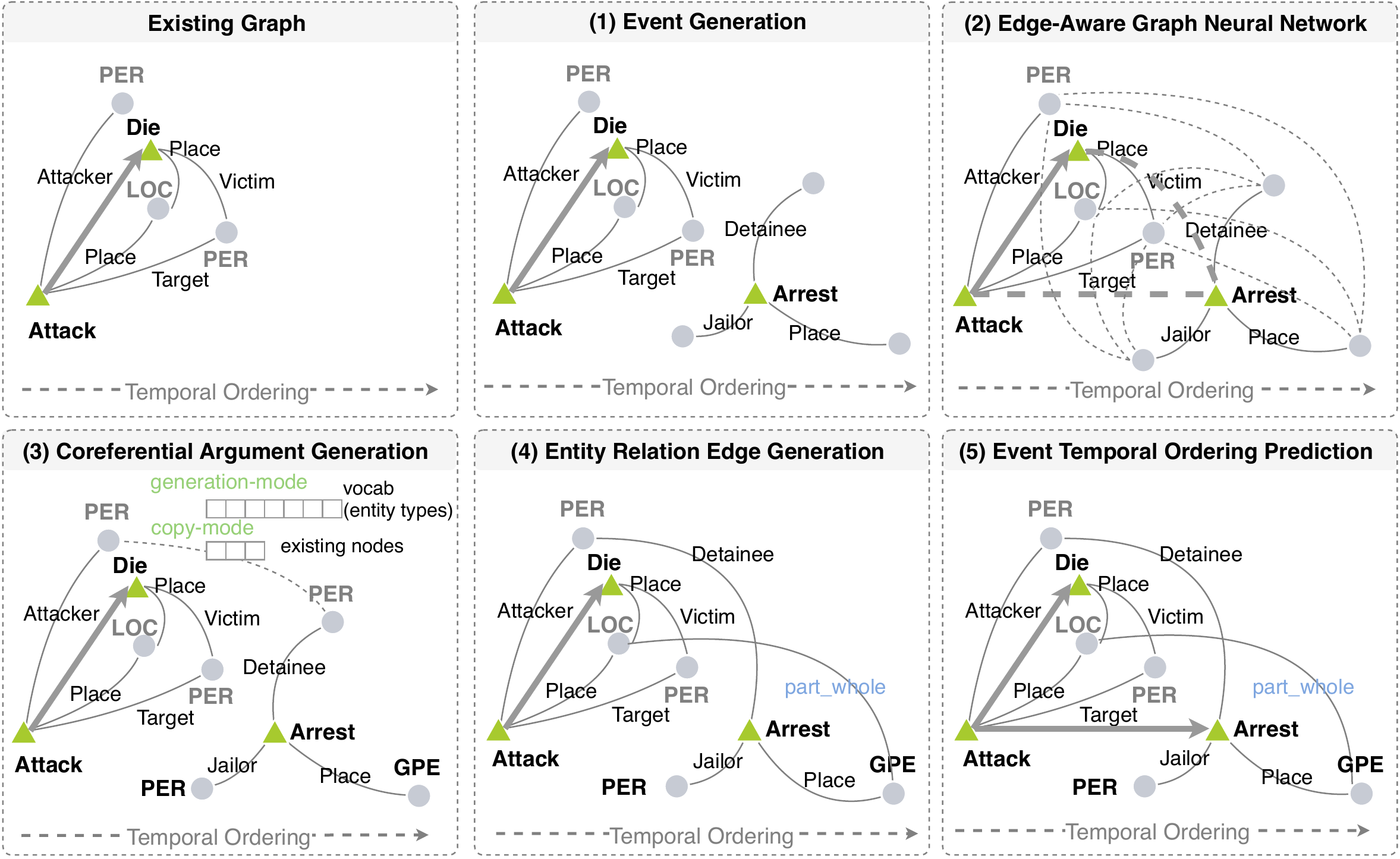}
    \caption{The generation process of Temporal Event Graph Model. 
    }
    \label{fig:overview}
\end{figure*}

\section{Problem Formulation}
\begin{table}[t]
\small
    \centering
    \begin{tabular}{c|m{17em}}
    \toprule 
       Symbol  & Meaning \\
       \midrule
 $G \in \mathcal{G}$         & Instance graph of a complex event \\
 $S \in \mathcal{S}$         & Schema graph of a complex event type \\
    $e \in \mathcal{E}$         & Event node in an instance  graph \\
    $v \in \mathcal{V}$         & Entity node in an instance graph \\
    $\langle e_i, e_l \rangle$         & Temporal ordering edge between events $e_i$ and $e_l$, indicating $e_i$ is before $e_l$ \\
    $\langle e_i, a, v_j \rangle$         & Argument edge, indicating $v_j$ plays argument role $a$ in the event $e_i$ \\
    $\langle v_j, r, v_k \rangle$         & Relation edge between entities $v_j$ and $v_k$, and $r$ is the relation type \\
    $\mathcal{A}(e)$ & Argument role set of event $e$, defined by the IE ontology \\
    $\Phi_{\mathcal{E}}$ & The type set of events \\
    $\Phi_{\mathcal{V}}$ & The type set of entities \\
    $\phi(\cdot)$ & A mapping function from a node to its type \\
    $G_{<i}$ & Subgraph of $G$ containing events before $e_i$ and their arguments \\
    \bottomrule
    \end{tabular}
    \caption{List of symbols 
    }
    \label{tab:symbols}
\end{table}

From a set of documents describing a complex event, we construct an \textbf{instance graph} $G$ which contains event nodes $E$ and entity nodes (argument nodes) $V$.
There are three types of edges in this graph: (1) event-event edges $\langle e_i, e_l \rangle$ connecting events that have direct temporal relations; (2) event-entity edges $\langle e_i, a, v_j \rangle$ connecting arguments to the event; and (3) entity-entity edges $\langle v_j, r, v_k \rangle$ indicating relations between entities. 
We can construct instance graphs by applying Information Extraction (IE) techniques on an input text corpus. %
In these graphs, the relation edges do not have directions but temporal edges between events are directional, going from the event before to the event after. 

For each complex event type, given a set of instance graphs $\mathcal{G}$, the goal of schema induction is to generate a schema library $\mathcal{S}$. In each \textbf{schema graph} $S$, the nodes are abstracted to the types of events and entities.
\figurename~\ref{fig:schema-example} is an example of schema\footnote{For simplification purposes, we mention ``{schema graphs}'' as ``{schemas}'', and ``events'' in schemas are only ``event types''. } %
for complex event type \textit{car-bombing}. Schema graphs can be regarded as a summary abstraction of instance graphs, capturing the reoccurring structures.   %

%% file: 3_method.tex
\section{Our Approach}

\input{3_0_instance_graph}

\input{3_1_overview}

\input{3_2_graphlm}

%% file: 3_0_instance_graph.tex
\subsection{Instance Graph Construction}
\label{sec:instance_graph_extraction}

\noindent
To induce schemas for a complex event type, such as \textit{car-bombing}, we construct a set of instance graphs, where each instance graph is about one complex event, such as \textit{Kabul ambulance bombing}. 

We first identify a cluster of documents that describes the same complex event. In this paper, we treat all documents linked to a single Wikipedia page as belonging to the same complex event, detailed in \S\ref{sec:dataset}. %

We use OneIE, a state-of-the-art Information Extraction system~\cite{lin2020joint}, to extract entities, relations and events, and then perform cross-document entity~\cite{pan2015unsupervised,pan2017cross} and event coreference resolution~\cite{Lai2021} over the document cluster of each complex event. We further conduct event-event temporal relation extraction~\cite{ning2019improved,Wen2021a} to determine the order of event pairs. 
{We run the entire pipeline following~\cite{wen2021resin}~\footnote{\url{https://github.com/RESIN-KAIROS/RESIN-pipeline-public}}, and the detailed extraction performance is reported in the paper.
}

{After extraction, we construct one instance graph for each complex event, where coreferential events or entities are merged. }
We consider the isolated events as irrelevant nodes in schema induction, so they are excluded from the instance graphs during graph construction. 
{Considering schema graphs focus on type-level abstraction, we use type label and node index to represent each node, ignoring the mention level information in these instance graphs.}

%% file: 3_1_overview.tex
\subsection{Temporal Event Graph Model Overview}

Given an instance graph $\mathcal{G}$
, we regard the schema as the hidden knowledge to guide the generation of these graphs. 
To this end, we propose a temporal event graph model that maximizes the probability of each instance graph, parameterized by $\prod_{G \in \mathcal{G}} p(G)$. 
At each step, based on the previous graph $G_{<i}$, we predict one event node $e_i$  with its arguments %
to generate the next graph $G_{i}$,  
\begin{equation*}
p(G) = \prod\nolimits_{i=0}^{|\mathcal{E}|} p(G_{i}|G_{<i}).
\end{equation*}
We factorize the probability of generating new nodes and edges as:
\begin{equation*}
\begin{aligned}
    & p(G_{i}|G_{<i}) = p(e_i| G_{<i}) 
    \!\!\!\! \prod_{a_j \in \mathcal{A}(e_i)} \!\!\!\!\! p(\langle e_i, a_j, v_j \rangle| e_i, a_j)  
    \\
    & \!  \prod_{v_k \in G_{<i}} \!\!\!\!  p(\langle v_j, r, v_k \rangle | v_j, v_k) 
    \!\!\!\!  \prod_{e_l \in G_{<i}} \!\!\!\! p(\langle e_i, e_l \rangle | e_i, e_l). 
    \ \ \ \text{(1)}
\end{aligned}
\end{equation*}
As shown in \figurename~\ref{fig:overview}, 
an event node $e_i$ is generated first according to the probability $p(e_i| G_{<i})$.
We then add argument nodes based on the IE ontology. %
We also predict relation $\langle v_j, r, v_k \rangle$ between the newly generated node $v_j$ and the existing nodes $v_k \in G_{<i}$.
After knowing the shared and related arguments, we add a final step to predict the temporal relations between the new event $e_i$ and the existing events $e_l \in G_{<i}$. 

In the traditional graph generation setting, the order of node generation can be arbitrary. However, in our instance graphs, event nodes are connected through temporal relations.
We order events as a directed acyclic graph (DAG). Considering each event may have multiple events both ``before'' and ``after'', we obtain the generation order by traversing the graph using Breadth-First Search. %

We also add dummy \textsc{Start}/\textsc{End} event nodes to indicate the starting/ending of the graph generation. %
At the beginning of the generation process, the graph $G_0$ has a single start event node $e_{\left[\text{SOG}\right]}$. %
We generate $e_{\left[\text{EOG}\right]}$ to signal the end of the graph. %

%% file: 3_2_graphlm.tex
\subsection{Event Generation}

To determine the event type of the newly generated event node $e_i$, we apply a graph pooling over all  events to get the current graph representation $\bs{g}_i$, 
\begin{equation*}
    \bs{g}_i =  \text{Pooling}(\{\bs{e}_0, \cdots, \bs{e}_{i-1}\}).
\end{equation*}
We use bold to denote the latent representations of nodes and edges, which will be initialized as zeros and updated at each generation step via message passing in \S~\ref{sec:gnn}. We adopt a mean-pooling operation in this paper. After that, the event type is predicted through a fully connected layer, 
\begin{equation*}
    p(e_i | G_{<i}) = \frac{\exp( \bs{W}_{\phi(e_i)} \bs{g}_i )}{\sum_{\phi' \in \Phi_{\mathcal{E}} \cup \left[\text{EOG}\right] } \exp( \bs{W}_{\phi'} \bs{g}_i )}. %
\end{equation*}

Once we know the event type of $e_i$, we add all of its arguments in $\mathcal{A}(e_i)$ defined in the IE ontology as new entity nodes.
For example, in \figurename~\ref{fig:overview}, the new event $e_i$ is an \textsc{Arrest} event, so we add three argument nodes for \textsc{Detainee}, \textsc{Jailor}, and \textsc{Place} respectively. 
The edges between these arguments and event $e_i$ are also added into the graph. %

\subsection{Edge-Aware Graph Neural Network}
\label{sec:gnn}

We use a Graph Neural Network (GNN)~\cite{kipf2016semi} to update node embeddings following the graph structure. %
Before we run the GNN on the graph, we first add \textit{virtual edges} between the newly generated event and all previous events, and between new entities and previous entities, shown as dashed lines in \figurename~\ref{fig:overview}. 
The virtual edges enable the representations of new nodes to aggregate the messages from previous nodes, which has been proven effective in~\cite{liao2019efficient}.

To capture rich semantics of edge types, we pass edge-aware messages during graph propagation. An intuitive way is to encode different edge types with different convolutional filters, which is similar to RGCN~\cite{schlichtkrull2018modeling}. However, the number of RGCN parameters grows rapidly with the number of edge types and easily becomes unmanageable given the large number of relation types and argument roles in the IE ontology.\footnote{There are 131 edge types according to the fine-grained LDC Schema Learning Ontology. %
} 
Instead, we learn a vector representation for each relation type $r$ and argument role $a$. 
The message passed through each argument edge $\langle e_i, a, v_j \rangle$ is: 
\begin{equation*}
    \bs{m}_{i,j} = \text{ReLU} \left (\bs{W}_a ( (\bs{e}_i - \bs{v}_j) \mathbin\Vert \bs{a} ) \right),
\end{equation*}
where $\mathbin\Vert$ denotes concatenation operation.
Similarly, the message between two entities $v_j$ and $v_k$ is: %
\begin{equation*}
    \bs{m}_{j,k} = \text{ReLU} \left (\bs{W}_r ( (\bs{v}_j - \bs{v}_k) \mathbin\Vert \bs{r} ) \right).
\end{equation*}
Considering that the direction of the temporal edge is important, we parametrize the message over this edge by assigning two separate weight matrices to the outgoing and incoming vertices: %
\begin{equation*}
    \bs{m}_{i,l} = \text{ReLU} \left  (\bs{W}_{\text{bfr}}\bs{e}_i - \bs{W}_{\text{aft}} \bs{e}_l   \right).
\end{equation*}

We aggregate the messages using edge-aware attention following~\cite{liao2019efficient}:\footnote{Compared to~\cite{liao2019efficient}, we do not use the positional embedding mask  because the newly generated nodes have distinct roles.}
\begin{equation*}
    \alpha_{i,j} = \sigma (\text{MLP} (\bs{e}_i - \bs{e_j})),
\end{equation*}
where $\sigma$ is the sigmoid function, and MLP contains two hidden layers with ReLU nonlinearities.

The event node representation $\bs{e}_i$ is then updated using the messages from its local neighbors $\mathcal{N}(e_i)$,  similar to entity node representations:
\[
    \bs{e}_i \leftarrow \text{GRU} \left(\bs{e}_i \mathbin\Vert \sum\nolimits_{j \in \mathcal{N}(e_i)} \alpha_{i,j} \bs{m}_{i,j}\right).
\]

\subsection{Coreferential Argument Generation}
\label{sec:coreferential}

After updating the node representations, we detect the entity type of each argument, and also predict whether the argument is coreferential to existing entities. 
Inspired by copy mechanism~\cite{gu2016incorporating}, we classify each argument node $v_j$ to either a new entity with entity type $\phi(v_j)$, or an existing entity node in the previous graph $G_{<i}$. For example, in \figurename~\ref{fig:overview}, the \textsc{Detainee} should be classified to the existing \textsc{Attacker} node, while \textsc{Jailor} node is classified as \textsc{Person}. Namely, 
\begin{equation*}
\begin{aligned}
    p(&\langle e_i, a_j, v_j \rangle| e_i, a_j)
    \\
    &=\begin{cases}
    p(\langle e_i, a_j, v_j \rangle, \texttt{g}| e_i, a_j)
    &\mathrm{if~} v_j \mathrm{~is~new},
    \\
    p(\langle e_i, a_j, v_j \rangle, \texttt{c}| e_i, a_j)
    &\mathrm{otherwise},
    \end{cases} 
\end{aligned}
\end{equation*}
where $p(\langle e_i, a_j, v_j \rangle, \texttt{g}| e_i, a_j)$ is the generation probability, classifying the new node to its entity type $\phi(v_j)$: %
\begin{equation*}
p(\langle e_i, a_j, v_j \rangle, \texttt{g}| e_i, a_j) = \exp( \bs{W}_{\phi(v_j)} \bs{v}_j ) \big/ Z %
\end{equation*}
The copy probability $p(\langle e_i, a_j, v_j \rangle, \texttt{c}| e_i, a_j)$ %
selects the coreferential entity $v$ from the entities in existing graph, denoted by $V_{<i}$,
\begin{equation*}
p(\langle e_i, a_j, v_j \rangle, \texttt{c}| e_i, a_j) = \exp( \bs{W}_{v} \bs{v}_j ) \big/ Z
.  
\end{equation*}
Here, $Z$ is the shared normalization term, 
\begin{equation*}
Z = \sum\nolimits_{\phi' \in \Phi_{ \mathcal{V}} } \!\!\!\! \exp( \bs{W}_{\phi'} \bs{v}_j ) + \sum\nolimits_{v' \in V_{<i}} \!\!\!\! \exp( \bs{W}_{v'} \bs{v}_j )
\end{equation*}
If determined to copy, we merge coreferential entities in the graph.

\subsection{Entity Relational Edge Generation}
\label{sec:relation}

In this phase, we determine the \textit{virtual edges} to be kept and assign relation types to them, such as \textsc{PartWhole} relation in \figurename~\ref{fig:overview}. We model the relation edge generation probability as a categorical distribution over relation types, and add $\left[\text{O}\right]$ (\textsc{Other}) to the typeset $\mathcal{R}$ to represent that there is no relation edge:
\begin{equation*}
\begin{aligned}
    p(\langle v_j, r, v_k \rangle | v_j, v_k) = \frac{\exp( \text{MLP}_r (\bs{v}_j - \bs{v}_k) )}
    { \!\!\!\!\! \sum\limits_{r' \in \mathcal{R} \cup \left[\text{O}\right]} \!\!\!\!\! \exp( \text{MLP}_{r'} (\bs{v}_j - \bs{v}_k) )}
\end{aligned}
\end{equation*}
We use two hidden layers with ReLU activation functions to implement the MLP.

\subsection{Event Temporal Ordering Prediction}
\label{sec:ordering}

To predict the temporal dependencies between the new events and existing events, we connect them through temporal edges, as shown in \figurename~\ref{fig:overview}. These edges are critical for message passing in predicting the next event. We build temporal edges in the last phase of generation, since it relies on the shared and related arguments.
Considering that temporal edges are interdependent, we model the generation probability as a mixture of Bernoulli distributions following~\cite{liao2019efficient}:
\begin{equation*}
\begin{aligned}
    & p(\langle e_i, e_l \rangle | e_i, e_l) = \sum\nolimits_b \gamma_b \ \theta_{b,i,l}, \\
    & \gamma_1, \cdots, \gamma_B = \text{Softmax} \left( \sum\nolimits_{i,l} \text{MLP} (\bs{e}_i - \bs{e}_l) \right), \\
    &\theta_{1,i,l}, \cdots, \theta_{B,i,l} = \sigma \left ( \text{MLP}_\theta (\bs{e}_i - \bs{e}_l) \right ),
\end{aligned}
\end{equation*}
where $B$ is the number of mixture components.  When $B=1$, the distribution degenerates to factorized Bernoulli, which assumes the independence of each potential temporal edge conditioned on the existing graph.

\subsection{Training and Schema Decoding}
We train the model by optimizing the negative log-likelihood loss, 
\begin{equation*}
    \mathcal{L} = 
    \sum\nolimits_{G \in \mathcal{G}_{\text{train}} } 
     -\log_{2} p(G) 
    .
\end{equation*}
To compose the schema library for each complex event scenario, we construct instance graphs from related documents to learn a graph model, and then obtain the schema using greedy decoding. %

%% file: 4_benchmark.tex
\section{Evaluation Benchmark}

\input{4_1_dataset}

\input{4_2_evaluation}

\input{4_3_instance_coverage}

\input{4_4_event_pred}

%% file: 4_1_dataset.tex
\subsection{Dataset}
\label{sec:dataset}

{We conduct experiments on two datasets for both the general scenario and a more specific scenario. We adopt the DARPA KAIROS\footnote{
 \url{https://github.com/NextCenturyCorporation/kairos-pub/tree/master/data-format/ontology}}
ontology, a newly defined fine-grained ontology for Schema Learning, with 24 entity types, 46 relation types, 67 event types, and 85 argument roles. \footnote{The ontology has been released in LDC2020E25. %
}}
{Our schema induction method does not rely on any specific ontology, only the IE system is trained on a given ontology to create the instance event graphs.}

\noindent
\textbf{General Schema Learning Corpus:} The Schema Learning Corpus, released by LDC (LDC2020E25), includes 82 types of complex events, such as \textit{Disease Outbreak}, \textit{Presentations} and \textit{Shop Online}. Each complex event is associated with a set of source documents. 
This data set also includes ground-truth schemas created by LDC annotators, which were %
used for our intrinsic evaluation.

\input{table/dataset}

\noindent
\textbf{IED Schema Learning Corpus:} The same type of complex events may have many variants, which depends on the different types of conditions and participants. In order to evaluate our model's capability at capturing uncertainty and multiple hypotheses, we decided to dive deeper into one scenario and chose the \textit{improvised explosive device (IED)} as our case study. 
We first collected Wikipedia articles that describe 4 types of complex events, i.e., \textit{Car-bombing IED}, \textit{Drone Strikes IED}, %
\textit{Suicide IED} and \textit{General IED}. Then we followed~\cite{li2021document} to exploit the external links to collect the additional news documents with the corresponding complex event type.

The ground-truth schemas for this IED corpus are created manually, through a schema curation tool~\cite{Mishra2021}. 
Only one human schema graph was created for each complex event type%
, resulting in 4 schemas. 
In detail, for each complex event type, we presented example instance graphs and the ranked event sequences to annotators to create human (ground truth) schemas. The event sequences are generated by traversing the instance graphs, and then sorted by frequency and the number of arguments. 
Initially we assigned three annotators (IE experts) to each create a version of the schema and then the final schema was merged through discussion. 
After that, two annotators (linguists) performed a two-pass revision. 
Human curation focuses on merging and trimming steps by validating them using the reference instance graphs. Also, temporal dependencies between steps were further refined, and coreferential entities and their relations were  added during the curation process.
To avoid bias from the event sequences, linguists in the second round revision were not presented with the event sequences. All annotators were trained and disagreements were resolved through discussion.

%% file: table/dataset.tex
\begin{table}[!hbt]
\small
\centering
\setlength\tabcolsep{3pt}
\begin{tabular}{c c ccccc}
\toprule
\textbf{Dataset} & \textbf{Split} & \textbf{\#doc}  & \textbf{\#graph}& \textbf{\#event} & \textbf{\#arg} & \textbf{\#rel} \\
\midrule
 &\textbf{Train} & 451 & 451  & 6,040 & 10,720 & 6,858  \\
 \textbf{General} & \textbf{Dev}   & 83  & 83 & 1,044  & 1,762 & 1,112    \\
 &\textbf{Test} & 83 & 83 & 1,211  & 2,112 & 1,363 \\
 \midrule
 &\textbf{Train} & 5,247 & 343 & 41,672 & 136,894 & 122,846  \\
 \textbf{IED}  &\textbf{Dev} & 575 & 42  & 4,661  & 15,404 & 13,320    \\
 &\textbf{Test} & 577  & 45 & 5,089  & 16,721 & 14,054 \\
\bottomrule
\end{tabular}
\caption{Data statistics. %
Each instance graph is about one complex event. }
\label{table:data_stats}
\end{table}

%% file: 4_2_evaluation.tex
\subsection{Schema Matching Evaluation} 
\label{sec:schema_matching_evaluation}

We compare the generated schemas with the ground truth schemas based on the overlap between them.
The following evaluation metrics were employed:\footnote{We cannot use graph matching to compare between baselines and our approach due to the difference in the graph structures being modeled.} %

\noindent
\textbf{Event Match:} A good schema must contain the  events crucial to the complex event scenario.   \textit{F-score} is used to compute the overlap of event nodes.

\noindent
\textbf{Event Sequence Match:} A good schema is able to track events through a timeline. So we obtain event sequences following temporal order, and evaluate \textit{F-score} on the overlapping sequences of lengths $l=2$ and $l=3$.

\noindent
 \textbf{Event Argument Connection Match:} Our complex event graph schema includes entities and their relations %
and captures how events are connected through arguments, in addition to their temporal order. %
We categorize these connections into three categories: (1) two events are connected by shared arguments; (2) two events have related arguments, i.e., their arguments are connected through entity relations; (3) there are no direct connections between two events. For every pair of overlapped events, we calculate \textit{F-score} based on whether these connections are predicted correctly. 
The human schemas of the General dataset do not contain arguments and the relations between arguments, so we only compute this metric for the IED dataset. 

%% file: 4_3_instance_coverage.tex
\subsection{Instance Graph Perplexity Evaluation}

To evaluate our temporal event graph model, we compute the \textit{instance graph perplexity} by predicting the instance graphs in the test set, 
\begin{equation}
    \mathrm{PP} = 2^{ -
    \frac{1}{|\mathcal{G}_{\text{test}}|} \sum\limits_{G \in \mathcal{G}_{\text{test}} } 
    \log_{2} p(G) 
    }.
\end{equation}
We calculate the \textit{full perplexity} for the entire graph using Equation (1), 
and \textit{event perplexity} using only event nodes, emphasizing the importance of correctly predicting events.

%% file: 4_4_event_pred.tex
\subsection{Schema-Guided Event Prediction}
\label{sec:ending_event_prediction}

To explore schema-guided probabilistic reasoning and prediction, we perform an extrinsic evaluation of event prediction. Different from traditional event prediction tasks, the temporal event graphs contain arguments with relations, and there are type labels assigned to nodes and edges. %
We create a graph-based event prediction dataset using our testing graphs. The task aims to predict ending events of each graph, i.e., events that have no future events after it.  An event is predicted correctly if its event type matches one of the ending events in the graph. Considering that there can be multiple ending events in one instance graph, we rank event type prediction scores and adopt \textit{MRR (Mean Reciprocal Rank)} and \textit{HITS@1} as evaluation metrics.

%% file: 5_analysis.tex
\input{table/schema_matching}

\section{Experiments}

\input{5_1_baseline}

\input{5_2_quantitative}

%% file: table/schema_matching.tex
\begin{table*}[!hbt]
\small
\centering
\setlength\tabcolsep{6pt}
\setlength\extrarowheight{2pt}
\begin{tabular}{l | l | c | c c  | c | c c} %
\toprule
\multirow{1}{*}{} 
& \multirow{3}{*}{\textbf{Model}} &
\textbf{\ \ \ \ Event \ \ \ \ } &
\multicolumn{2}{c|}{\textbf{Sequence Match}} &
\textbf{Connection}
 & \textbf{Event} & \textbf{Full}
\\ %
\multirow{1}{*}{\textbf{Dataset}} 
& 
& \textbf{ Match } 
& \textbf{ $l=2$ } & \textbf{ $l=3$ } %
& \textbf{Match} 
& \textbf{Perplexity} & \textbf{Perplexity}
\\
\midrule
\multirow{4}{*}{ {\!General}}

& {Event Language Model} 
& 54.76 & 22.87
& 8.61 %
& -
& - & -
\\
& {Sequential Pattern Mining}
& 49.18 & 20.31
& 7.37 %
& - 
& - & -
\\\cmidrule{2-8}
& \textbf{Event Graph Model}
& \textbf{58.15} & \textbf{24.79} 
& \textbf{9.18} %
& {-}
& \textbf{24.25} & 137.18
\\
& \ \ \  w/o ArgumentGeneration
& 56.96 & 22.47 
& 8.21 %
& -
&  68.59 & -
\\
\midrule
\multirow{4}{*}{ {\!IED}}
& {Event Language Model} 
& 49.15 & 17.77 
& 5.32 %
& -
& - & -
\\
& {Sequential Pattern Mining}
& 47.91 & 18.39
& 4.79 %
& 5.41
& - & -
\\\cmidrule{2-8}
& \textbf{Event Graph Model}
& \textbf{59.73} & \textbf{21.51}
& \textbf{7.81} %
& \textbf{10.67}
& \textbf{39.39} & 168.89
\\
& \ \ \  w/o ArgumentGeneration
& 55.01 & 18.24
& 6.67 %
& -
& 51.98 & - 
\\
\bottomrule
\end{tabular}
\caption{Intrinsic evaluation results, including schema matching F1 score (\%) and instance graph perplexity. %
}
\label{table:result_schema_matching}
\end{table*}

%% file: 5_1_baseline.tex
\subsection{Experiment Setting}

\textbf{Baseline 1: Event Language Model}~\cite{rudinger2015script,pichotta2016learning} is the state-of-the-art event schema induction method. It learns the probability of temporal event sequences, %
and the event sequences generated from event language model are considered as schemas. 

\noindent
\textbf{Baseline 2: Sequential Pattern Mining}~\cite{han2001prefixspan}
is a classic algorithm for discovering common sequences. 
We also attach arguments and their relations as extensions to the pattern. %
Considering that the event language model baseline cannot handle multiple arguments and relations, we add sequential pattern mining for comparison. The frequent patterns mined are considered as schemas.

\noindent
\textbf{Reference: Human Schema}
is added as a baseline in the extrinsic task of event prediction. %
Since human-created schemas are highly accurate but not probabilistic, we want to evaluate their limits at predicting events in the extrinsic task. 
We match schemas to instances and fill in the matched type. %

\noindent
\textbf{Ablation Study: Event Graph Model w/o Argument Generation} is included as a variant of our %
model in which we remove argument generation (\S\ref{sec:coreferential} and~\S\ref{sec:relation}). It learns to generate a graph containing only event nodes with their temporal relations, %
aiming to verify whether incorporating argument information helps event modeling.

\subsection{Implementation Details} 

\noindent
\textbf{Training Details}. 
{
For our event graph model, the representation dimension is 128, and we use a 2-layer GNN. The value of $B$ is 2. The number of mixture components in temporal classifier is 2. The learning rate is 1e-4. %
To train event language model baseline, instead of using LSTM-based architecture following~\cite{pichotta2016learning}, we adopt the state-of-the-art auto-regressive language XLNet~\citep{yang2019xlnet}. In detail, we first linearize the graph using topological sort, and then train XLNet\footnote{\url{https://github.com/huggingface}}  using the dimension of 128 (the same as our temporal event graph model), and the number of layers is 3. The learning rate is 1e-4. We select the best model on the validation set. Both of our model and event language model baseline are trained on one Tesla V100 GPU with 16GB DRAM.
For sequential pattern mining, we perform random walk, starting from every node in instance graphs and ending at sink nodes, to obtain event type sequences, and then apply PrefixSpan~\cite{han2001prefixspan} \footnote{\url{https://github.com/chuanconggao/PrefixSpan-py}} 
to rank sequential patterns. %
}

\noindent
\textbf{Evaluation Details.}
To compose the schema library, %
we use the first ranked sequence as the schema for these two models. %
To perform event prediction using baselines, we traverse the input graph to obtain event type sequences, and conduct prediction on all sequences to produce an averaged score. For human schemas, we first linearize them and the input graphs, and find the longest common subsequence between them. %

%% file: 5_2_quantitative.tex
\input{table/event_pred}

\subsection{Results and Analysis}

\noindent
\textbf{Intrinsic Evaluation}.
In \tablename~\ref{table:result_schema_matching}, the significant gain on \textit{event match} %
demonstrates the ability of our graph model to keep salient events. %
On \textit{sequence match}, our approach achieves larger performance gain compared to baselines when the path length $l$ is longer. It implies that the proposed model is capable of capturing longer and wider temporal dependencies.
In the case of \textit{connection match}, only sequential pattern mining in the baselines can predict connections between events. When compared against sequential pattern mining, our generation model significantly performs better since it considers the inter-dependency of arguments and encodes them with graph structures.

\begin{figure}[h]
    \centering
    \includegraphics[width=1.0\linewidth]{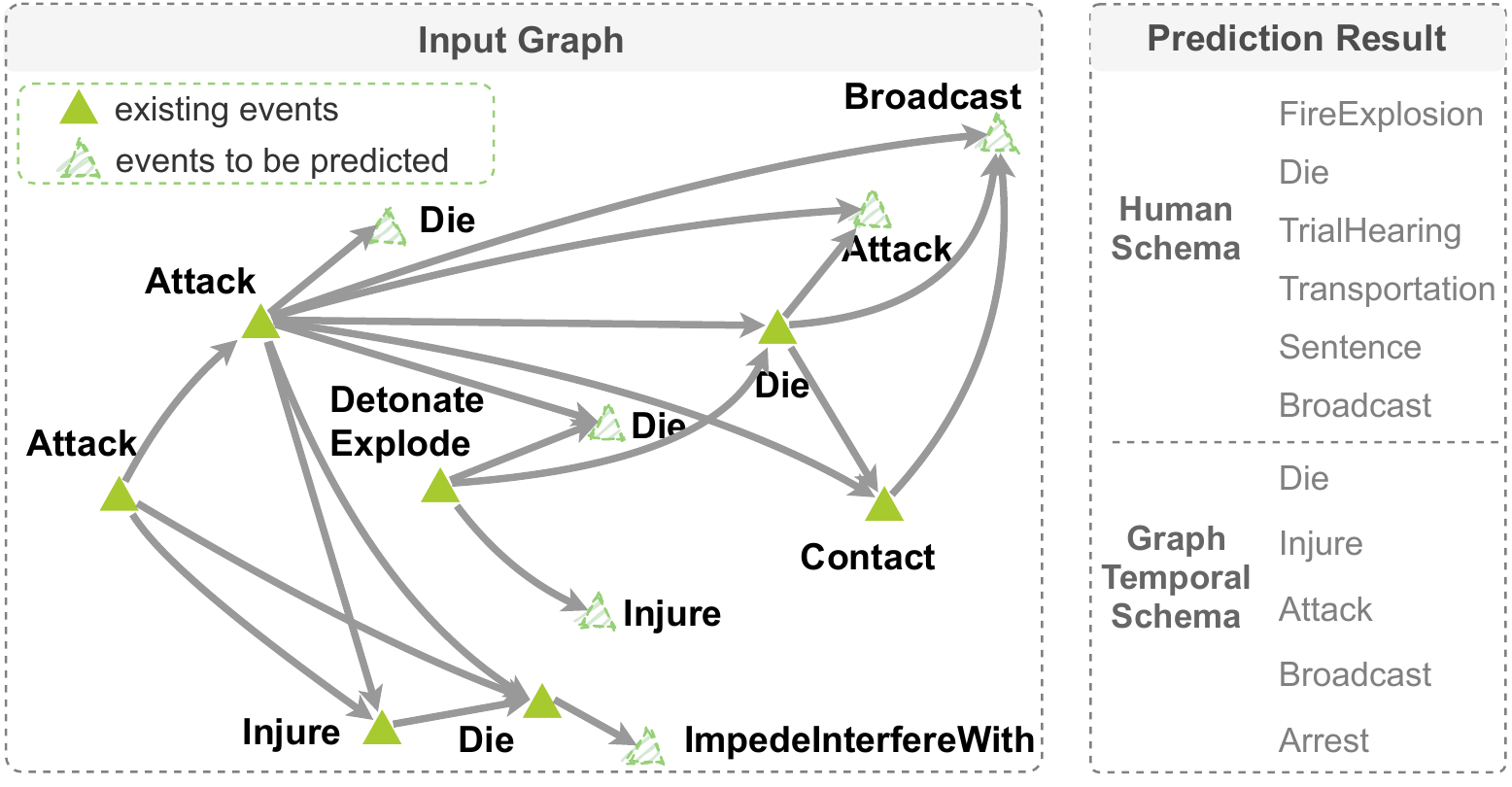}
    \caption{An event prediction example (IED scenario). 
    }
    \label{fig:example_pred}
\end{figure}

\noindent
\textbf{Extrinsic Evaluation}. 
On the task of schema-guided event prediction, our graph model obtains significant improvement (see \tablename~\ref{table:result_event_pred}.) The low performance of human schema demonstrates the importance of probabilistically modeling schemas to support downstream tasks. 
Take~\figurename~\ref{fig:example_pred} as an example. Human schemas produce incorrect event types such as \textsc{TrailHearing}, since it matches the sequence {\small \textsc{Attack}$\rightarrow$\textsc{Die}$\rightarrow$\textsc{TrailHearing}}, incapable of capturing the inter-dependencies between sequences. %
However, our model is able to customize the prediction to the global context of the input graph, and take into account that there is no \textsc{Arrest} event or justice-related events in the input graph. Also, the human schema fails to predict \textsc{Injure} and \textsc{Attack}, because it relies on the exact match of event sequences of lengths $l \geq 2$, and  cannot handle the variants of sequences. 
This problem can be solved by our probabilistic schema, via modeling the prediction probability conditioned on the existing graph. For example, even though \textsc{Attack} mostly happens before \textsc{Die}, we learn that \textsc{Attack} might repeat after \textsc{Die} event if there are multiple \textsc{Attack} and \textsc{Detonate} in the existing graph, which means the complex event is about a series of conflict events.  

\noindent
\textbf{Ablation Study}.
Removing argument generation (``w/o ArgumentGeneration'') generally lowers the performance on all evaluation tasks, since it ignores the coreferential arguments and their relations, but relies solely on the overly simplistic temporal order to connect events.  %
This is especially apparent from the instance graph perplexity in  \tablename~\ref{table:result_schema_matching}. %

\noindent
\textbf{Learning Corpus Size}. 
An average of 113 instance graphs is used for each complex event type in the IED scenario, %
and 383 instance graphs to learn the schema model in the General scenario. The better performance on the IED dataset in \tablename~\ref{table:result_schema_matching} %
shows that the number of instance graphs increases the schema induction performance. %

{
\noindent
\textbf{Effect of Information Extraction Errors.} Based on the error analysis for schemas induced in Table 1, the effect of extraction errors can be categorized into: (1) temporal ordering errors: 43.3\%; (2) missing events: 34.4\%; (3) missing coreferential events: 8.8\%; (4) incorrect event type: 7.7\%; (5) missing coreferential arguments: 5.5\%. However, even on automatically extracted event graphs with extraction errors, our model significantly performs better on event prediction compared to human-constructed schemas, as shown in Table 4. It demonstrates that our schema induction method is robust and effective to support downstream tasks, even when only provided with noisy data with extraction errors.
}

%% file: table/event_pred.tex
\begin{table}[!hbt]
\small
\centering
\setlength\tabcolsep{2pt}
\setlength\extrarowheight{2pt}
\begin{tabular}{l | l  | c | c  } %
\toprule
\textbf{Dataset} & \textbf{Model} & \textbf{ \ \  MRR \ \  } & \textbf{ HITS@1 } %
\\ 
\midrule
\multirow{6}{*}{{\!General}}

& {Event Language Model} 
& 0.367 & 0.497
\\
& {Sequential Pattern Mining}
& 0.330 & 0.478
\\
& {Human Schema} 
& 0.173 & 0.205
\\\cmidrule{2-4}
& \textbf{Event Graph Model}
& \textbf{0.401} & \textbf{0.520}
\\
& \ \ \  w/o ArgumentGeneraion
& 0.392 & 0.509
\\
\midrule
\multirow{6}{*}{{\!IED}}
& {Event Language Model} 
& 0.169 & 0.513
\\
& {Sequential Pattern Mining}
& 0.138 & 0.378
\\
& {Human Schema} 
& 0.072 & 0.222
\\\cmidrule{2-4}
& \textbf{Event Graph Model}
& \textbf{0.224} & \textbf{0.741}
\\
& \ \ \  w/o ArgumentGeneraion
& 0.210 & 0.734
\\
\bottomrule
\end{tabular}
\caption{Schema-guided event prediction performance. 
}
\label{table:result_event_pred}
\end{table}

%% file: 6_relatedwork.tex
\section{Related Work}

The definition of a \textit{complex event schema} separates us from related lines of work, namely \textit{schema induction} and \textit{script learning}. Previous work on \textit{schema induction} aims to characterize event triggers and participants of individual atomic events~\cite{chambers2013event,cheung2013probabilistic,nguyen2015generative, sha2016joint,yuan2018open}, ignoring inter-event relations.
Work on \textit{script learning}, on the other hand, originally limited attention to event chains with a single protagonist~\cite{chambers2008unsupervised, Chambers2009,rudinger2015script,jans2012skip,granroth2016happens} and later extended to multiple participants~\cite{pichotta2014statistical,pichotta2016learning,weber2017event}.
Recent efforts rely on distributed representations encoded from the compositional nature of events~\cite{modi2016event,granroth2016happens, weber2017event, weber2020causal, lyu-zhang-wikihow:2020}, and language modeling~\cite{rudinger2015script,pichotta2016learning,peng2016two}. %
All of these methods still assume that events follow linear order in a single chain. They also overlook the relations between participants which are critical for understanding the complex event. However, we induce a comprehensive event graph schema, capturing both the temporal dependency and the multi-hop argument dependency across events.

Our recent work on event graph schema induction~\cite{Li2020c} only considers the connections between a pair of two events. 
Similarly, their event prediction task is designed to automatically generate a missing event (e.g., a word sequence) given a single or a sequence of prerequisite events~\cite{nguyen2017sequence,hu2017happens,li2018constructing,kiyomaru2019diversity,lv2019sam}, or predict a pre-condition event given the current events~\cite{kwon-etal-2020-modeling}. 
In contrast, we leverage the automatically discovered temporal event schema as guidance to forecast the future events.

{
Existing script annotations~\cite{chambers2008unsupervised, chambers2010database, modi2017inscript, wanzare2016crowdsourced, mostafazadeh2016corpus, mostafazadeh2016caters, kwon-etal-2020-modeling} cannot support a comprehensive graph schema induction due to the missing of critical event graph structures, such as argument relations. Furthermore, in real-world applications, complex event schemas are expected to be induced from large-scale historical data, which is not feasible to annotate manually. We propose a data-driven schema induction approach, and choose to use IE systems instead of using manual annotation, to induce schemas that are robust and can tolerate extraction errors. 
}

Our work is also related to recent advances in modeling and generation of graphs~\cite{li2018learning,jin2018junction,grover2019graphite,simonovsky2018graphvae,liu2019graph,fu2020local,dai2020scalable, you2018graphrnn, liao2019efficient, yoo2020graph, shi2019graphaf}. %
We are the first to perform graph generation on event graphs.

%% file: 7_conclusion.tex
\section{Conclusions and Future Work}

We propose a new task to induce \textit{temporal complex event schemas}, which are capable of representing multiple temporal dependencies between events and their connected arguments. We induce such schemas by learning an \textit{event graph model}, a deep auto-regressive model, from the automatically extracted instance graphs. 
Experiments demonstrate the model's effectiveness on both intrinsic evaluation and the downstream task of schema-guided event prediction. These schemas can guide our understanding and ability to make predictions with respect to what might happen next, along with background knowledge including location-, and participant-specific and temporally ordered event information.  In the future, we plan to extend our framework to hierarchical event schema induction, as well as event and argument instance prediction.